\documentclass[10pt,twocolumn,letterpaper]{article}

\usepackage[pagenumbers]{iccv} % To force page numbers, e.g. for an arXiv version

\newcommand{\red}[1]{{\color{red}#1}}
\definecolor{darkyellow}{HTML}{CC9900}
\newcommand{\yellow}[1]{{\color{darkyellow}#1}}

\usepackage{graphicx}
\usepackage{amsmath}
\usepackage{amssymb}
\usepackage{booktabs}
\usepackage{multirow}
\usepackage{rotating}
\usepackage[normalem]{ulem}
\useunder{\uline}{\ul}{}
\usepackage{epigraph} 
\usepackage{dsfont}
\usepackage{bm}
\usepackage{enumitem} %
\usepackage[accsupp]{axessibility}

\usepackage{pifont}%

\newcommand{\nerfmatch}{NeRFMatch\xspace}

\definecolor{cvprblue}{rgb}{0.21,0.49,0.74}
\newcommand{\PAR}[1]{\noindent{\bf #1~}}

\usepackage{float}
\usepackage{placeins}

\newcommand{\pnp}{PnP\xspace}
\newcommand{\method}{GSVisLoc\xspace}
\newcommand{\gs}{GS\xspace}
\newcommand{\mastr}{MASt3R\xspace}
\newcommand{\gscpr}{GS-CPR\xspace}
\newcommand{\ransac}{RANSAC\xspace}
\newcommand{\sixdgs}{6DGS\xspace}
\newcommand{\scannetp}{ScanNet++ V2\xspace}

\usepackage{array}

\usepackage{adjustbox}

\definecolor{iccvblue}{rgb}{0.21,0.49,0.74}
\usepackage[pagebackref,breaklinks,colorlinks,allcolors=iccvblue]{hyperref}

\def\paperID{10} % *** Enter the Paper ID here
\def\confName{ICCV}
\def\confYear{2025}

\title{GSVisLoc: Generalizable Visual Localization for Gaussian Splatting Scene Representations}

\author{
Fadi Khatib$^{1*}$,
Dror Moran$^{1*}$,
Guy Trostianetsky$^{1}$,
Yoni Kasten$^{2}$,
Meirav Galun$^{1}$,
Ronen Basri$^{1}$ \\
\small $^{1}$Weizmann Institute of Science \quad
$^{2}$NVIDIA \\
\small Project Webpage: \url{https://gsvisloc.github.io/}
}

\begin{document}

 \maketitle
 {\let\thefootnote\relax\footnotetext{* The authors contributed equally and are listed alphabetically.
}}

 \begin{abstract}
We introduce GSVisLoc, a visual localization method designed for 3D Gaussian Splatting (3DGS) scene representations. Given a 3DGS model of a scene and a query image, our goal is to estimate the camera's position and orientation. We accomplish this by robustly matching scene features to image features. Scene features are produced by downsampling and encoding the 3D Gaussians while image features are obtained by encoding image patches. Our algorithm proceeds in three steps, starting with coarse matching, then fine matching, and finally by applying pose refinement for an accurate final estimate. Importantly, our method leverages the explicit 3DGS scene representation for visual localization without requiring modifications, retraining, or additional reference images. We evaluate GSVisLoc on both indoor and outdoor scenes, demonstrating competitive localization performance on standard benchmarks while outperforming existing 3DGS-based baselines. Moreover, our approach generalizes effectively to novel scenes without additional training.
\end{abstract}
 
 \section{Introduction}
\label{sec:intro}

Visual localization is the problem of estimating the camera position and orientation in a 3D environment given a query image. Solving the localization problem is crucial in many applications such as autonomous driving \cite{heng2019project}, robot navigation \cite{wendel2011natural}, and augmented reality \cite{ventura2014global}.

Visual localization approaches can be categorized by the underlying scene representation; see review in Section~\ref{sec:related work}. Classical, structure-based methods \cite{taira2018inloc, li2010location, camposeco2019hybrid, sattler2016efficient, irschara2009structure, sarlin2019coarse} rely on explicit 3D models with keypoint descriptors, achieving accurate localization by matching 2D query image pixels to 3D model points. 
End-to-end learned approaches implicitly encode the scene through the weights of a neural network. These methods include absolute pose regression (APR) \cite{kendall2015posenet, walch2017image, kendall2017geometric, brahmbhatt2018geometry, blanton2020extending, shavit2021learning} and scene coordinate regression (SCR) methods \cite{shotton2013scene, brachmann2017dsac, brachmann2021visual, brachmann2023accelerated, li2020hierarchical, yang2019sanet, brachmann2018learning, wang2024glace}. 
These learned implicit representations are optimized exclusively for visual localization. \cite{yen2021inerf, chen2024neural} use NeRF models \cite{mildenhall2020nerf} to refine the camera pose at test time through iterative rendering and pose adjustments, while 
\cite{zhou2024nerfect, liu2023nerf, moreau2023crossfire} work toward directly estimating 3D-2D correspondences from NeRF.

Recently, 3D Gaussian Splatting (3DGS) \cite{kerbl3Dgaussians} has gained attention as a promising scene representation for novel-view synthesis (NVS), offering fast training times and high-quality real-time rendering. Its growing popularity has spurred various studies exploring fundamental computer vision tasks within the 3DGS framework, including 3D segmentation \cite{cen2023segment}, registration \cite{chang2024gaussreg}, and 3D editing \cite{chen2024gaussianeditor}.
Several methods have been proposed to address the localization challenge for 3DGS. However, they either only refine an initial pose \cite{sun2023icomma, trivigno2024unreasonable, liu2025gscpr}, exhibit lower accuracy compared to other scene representations (6DGS \cite{bortolon20246dgs}), or require optimizing a specific 3DGS model (GSplatLoc \cite{sidorov2024gsplatloc}).

\begin{figure*}[t]
    \centering
    \includegraphics[width=0.7\linewidth, trim={0cm 0cm 0cm 0cm}]{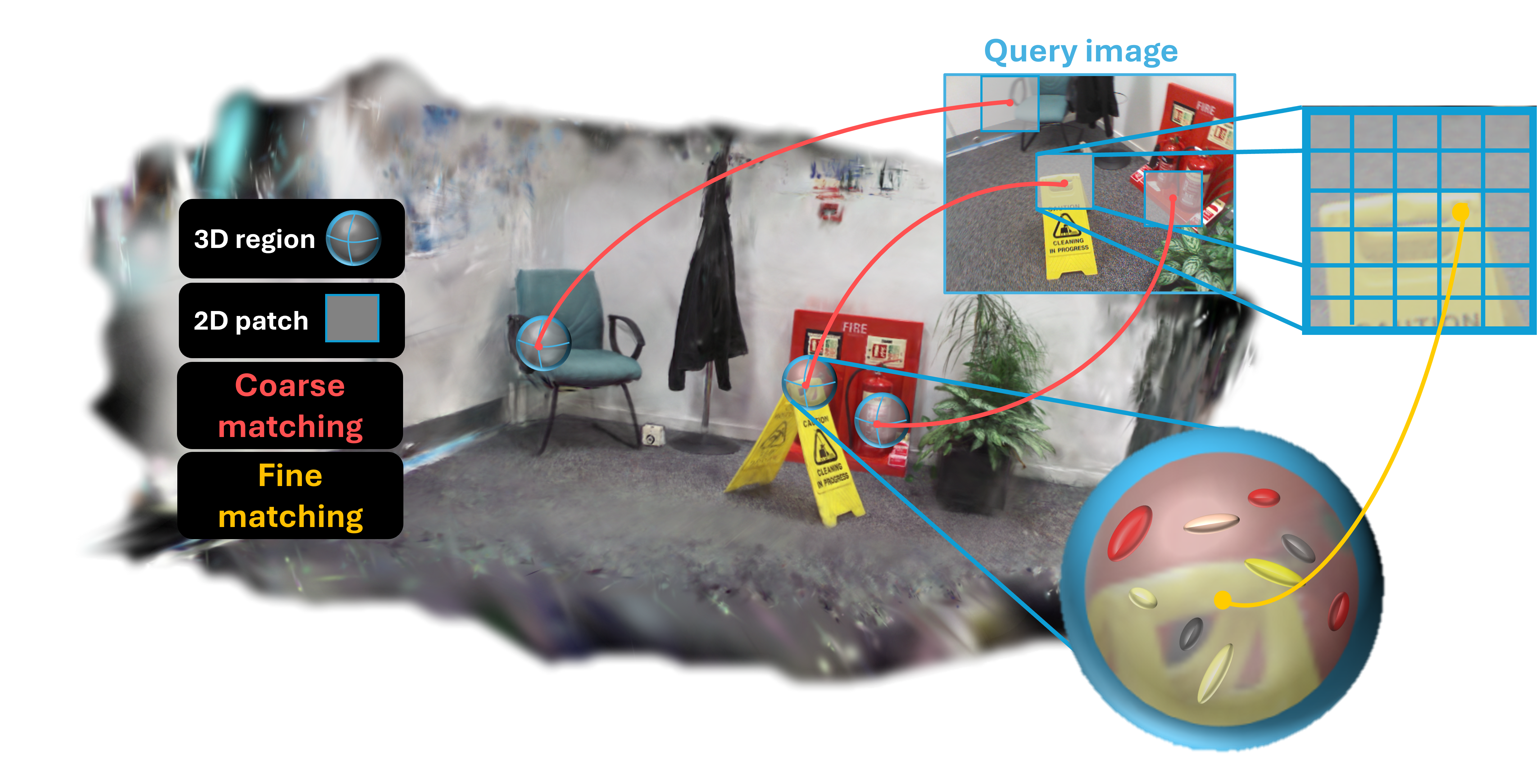}
    \caption{\textbf{\method}. Our method estimates coarse 3D–2D matches (shown in \red{red} in the figure) between 3D regions in the 3DGS and patches in the query image, and then refines them to pixel-level correspondences (\yellow{yellow}). These fine 3D–2D matches are fed into a PnP + RANSAC pipeline, followed by a pose refinement module, to produce the final camera pose of the query image.}
    \label{fig:OurModel}
\end{figure*}

In this paper, we propose to leverage the explicit scene representation of 3DGS for visual localization. Specifically, we introduce GSVisLoc, a generalizable deep-learning-based network for visual localization by matching 3DGS with 2D image features (see Figure~\ref{fig:OurModel}). Given a query image and a 3D Gaussian Splatting scene representation, we use a KPConv-based encoder \cite{thomas2019kpconv} to encode and downsample the Gaussians, producing features representing regions in 3D space. Additionally, we use an image encoder to extract features representing image patches from the query image. We then establish 3D-2D correspondences between the features, first at a coarse scale and then at a fine scale, enabling us to obtain an estimation for the camera pose. This is followed by an additional pose refinement step to achieve the final camera pose prediction.

Our method achieves accurate pose estimation that surpasses previous 3DGS-based localization methods. Moreover, \method's accuracy is competitive with state-of-the-art baselines on the 7-scenes dataset \cite{glocker2013real, shotton2013scene}. Notably, our model generalizes to novel, unseen scenes without requiring any modifications or retraining for the 3DGS representation, and it eliminates the need for an image retrieval step.
\newline

\noindent
In summary, our contributions include:
\begin{enumerate}
    
    \item We introduce \method, a deep neural network for visual localization for scenes represented using 3D Gaussian Splatting (3DGS).
    
    \item Our approach requires no retraining or modifications to the underlying 3DGS representation, simplifying its deployment.
    
    \item In contrast to structure-based and NeRF-based approaches, \method eliminates the need for image retrieval during inference, allowing reference images to be discarded after training.
    
    \item \method achieves state-of-the-art performance among 3DGS-based methods across all evaluated datasets, with accuracy competitive with state-of-the-art methods on the 7-scenes dataset.
    
    \item Most importantly, \method generalizes to novel, unseen scenes through learned 3D-2D matching, retaining accuracy comparable to single-scene trained models.

\end{enumerate}
 \section{Related work}
\label{sec:related work}

Visual localization approaches can be grouped according to their underlying scene representation:\\

\PAR{Structure-based Localization.}Structure-based localization methods \cite{taira2018inloc, li2010location, camposeco2019hybrid, sattler2016efficient, irschara2009structure, sarlin2019coarse} operate by establishing 3D-2D correspondences between 3D scene points and features in the query image. Camera pose is then calculated using a Perspective-n-Point (PnP) solver \cite{ke2017efficient, gao2003complete, kneip2011novel}. To streamline the matching process, an initial image retrieval step is used \cite{torii201524, hausler2021patch, arandjelovic2016netvlad, berton2022rethinking} to coarsely localize the query image, effectively narrowing the search space \cite{wang2024dgc, zhou2022geometry, li20232d3d}.

While this approach achieves accurate results, it requires substantial storage resources. Visual features \cite{dusmanu2019d2, detone2018superpoint, sarlin2020superglue, wang2020learning, zhou2021patch2pix, sun2021loftr, chen2022aspanformer} are typically extracted from a database of scene images to represent 3D points, and these features are then matched to features extracted from the query image to establish 3D-2D correspondences. To expedite inference, 3D descriptors are precomputed and stored alongside the scene model. This design, however, can lead to large memory footprints, complicating map updates and scalability \cite{zhou2022geometry}. Recent work has shown that storage requirements can be alleviated through compression and descriptor quantization \cite{wang2024mad, laskar2024differentiable}. Complementary to compression, descriptor-free 2D--3D matching establishes pixel-to-point correspondences via learned geometric and photometric reasoning, avoiding persistent high-dimensional 3D descriptors \cite{zhou2022geometry, li20232d3d, wang2024dgc}.

\PAR{End-to-End Learned Localization.}Absolute Pose Regression (APR) methods \cite{kendall2015posenet, walch2017image, kendall2017geometric, brahmbhatt2018geometry, blanton2020extending, shavit2021learning} train models to directly predict the camera pose from a query image, eliminating the need for explicit 3D representations. Although APR methods offer high-speed performance, they typically lack the accuracy and generalization capabilities of structure-based approaches~\cite{sattler2019understanding}.

Relative Pose Regression (RPR) methods predict the transformation between image pairs instead of the absolute pose of a single query. This often improves generalization over APR~\cite{balntas2018relocnet, khatib2024leveraging, arnold2022map, melekhov2017relative, en2018rpnet}, but RPR alone still falls short of the accuracy achieved by structure-based approaches.

Scene Coordinate Regression (SCR) methods \cite{shotton2013scene, brachmann2017dsac, brachmann2021visual, brachmann2023accelerated, li2020hierarchical, yang2019sanet, brachmann2018learning, wang2024glace} perform implicit 3D-2D matching by regressing the 3D scene coordinates directly from the query image. Like APR, these approaches use network parameters to encode the geometry of the scene~\cite{brachmann2017dsac, brachmann2021visual, brachmann2023accelerated, li2020hierarchical, brachmann2018learning}, but they are limited in representing large-scale scenes due to memory constraints. To address this limitation, more recent scene-agnostic SCR methods \cite{yang2019sanet, tang2023neumap} have decoupled the scene representation from the learned matching function, achieving scalability to larger scenes.

\PAR{NeRF-based Pose Estimation.}
NeRF-based pose estimation methods \cite{chen2024neural, yen2021inerf, lin2023parallel, pietrantoni2024self} rely on iterative rendering and pose adjustments, resulting in slow convergence and limited accuracy. NeFeS \cite{chen2024neural} improves APR pose estimation but struggles with SCR enhancements and suffers from a lengthy refinement time. HR-APR \cite{liu2024hr} accelerates the optimization process, yet each query still takes several seconds, even on a high-performance GPU. Other NeRF-based approaches, such as FQN \cite{germain2022feature}, CrossFire \cite{moreau2023crossfire}, NeRFLoc \cite{liu2023nerf}, and NeRFMatch \cite{zhou2024nerfect}, enhance positioning accuracy by establishing 3D-2D correspondences. However, these methods require an image retrieval step during inference.

\PAR{3DGS-based Pose Estimation.} Following the recent introduction of 3D Gaussian-Splatting (3DGS) for novel view synthesis (NVS), several methods have explored its use also for pose estimation. \sixdgs \cite{bortolon20246dgs}, which is closely related to our approach, achieves one-shot pose estimation by projecting rays from an ellipsoid surface, thus avoiding iterative processing. Although \sixdgs uses 3DGS for visual localization, it achieves inferior accuracies compared to previous methods. GSplatLoc \cite{sidorov2024gsplatloc}, a concurrent work, trains a Feature-3DGS model \cite{zhou2024feature} in which each Gaussian is assigned a feature through training the Gaussian splatting model. This approach establishes correspondences by matching 3D to 2D features through a mutual nearest-neighbor search. It then uses the \pnp algorithm to obtain an initial coarse pose estimate. The pose is subsequently refined iteratively to improve accuracy. In contrast to this method, our method uses the vanilla 3DGS model \cite{kerbl3Dgaussians} and does not require optimizing a specific Gaussian model. 

\PAR{3DGS-based Pose Refinement.}
Recent works have utilized 3DGS specifically for pose refinement. For example, iComMa \cite{sun2023icomma}, and MCLoc \cite{trivigno2024unreasonable}, inspired by iNeRF \cite{yen2021inerf}, implements an iterative refinement process for camera pose estimation by inverting 3DGS. Similarly, \gscpr\cite{liu2025gscpr} combines 3DGS with Mast3R \cite{leroy2024grounding}, a powerful image matcher, to apply pose refinement.

Our method establishes precise 3D-2D correspondences and seamlessly integrates GS-CPR’s pipeline as a plug-and-play module for pose refinement, achieving superior performance over other 3DGS-based methods.

 \begin{figure*}[!t]
    \centering
    \includegraphics[width=1.0\linewidth, trim={1cm 3cm 6cm 3cm}]{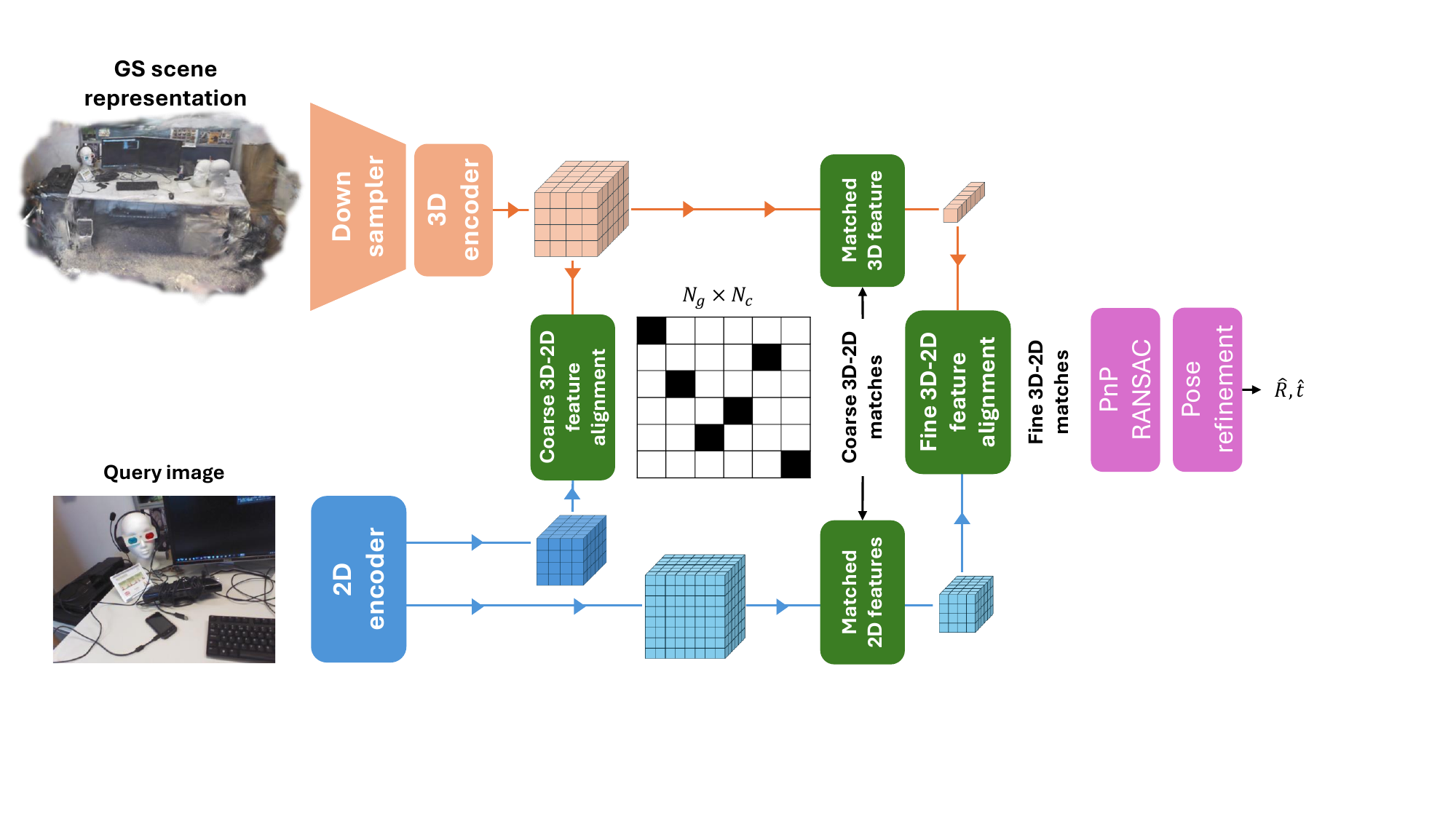}
    \caption{\textbf{\method architecture.} GSVisLoc uses a 3D Gaussian Splatting (3DGS) scene representation, processed by a 3D encoder, and a query image processed by a 2D encoder. Coarse 3D-2D matching establishes initial correspondences between the image and the 3D scene, which are refined to pixel-level matches. The final 3D-2D correspondences are passed through \pnp with \ransac for pose estimation, followed by a pose refinement step, yielding the final pose \((\hat{R}, \hat{t})\) of the query image. }
    \label{fig:model}
\end{figure*}

\section{Method}
\label{sec:method}
\subsection{Overview}
\label{subsec:overview}
We aim to establish point correspondences between a 3DGS scene representation and a 2D query image of the same scene. With these 3D-2D matches, a Perspective-n-Point (\pnp) solver \cite{ke2017efficient, gao2003complete, kneip2011novel} can efficiently estimate the camera pose.
To accomplish this, we propose \method, a deep neural network for visual localization. Our network encodes the 3D Gaussians and the query image in latent feature spaces. It then matches similar features and uses those matches to estimate the camera parameters. We further enhance our results by refining the initial pose estimate. Details of the architecture, training process, and pose refinement are provided in the following sections.

\subsection{\method}

Our network consists of a 3D\gs encoder, a 2D encoder, and cross-modal matching modules to align features from both domains and establish 3D-2D correspondences. The 2D encoder encodes two types of features for coarse and fine matching. Refer to Figure~\ref{fig:model} for a detailed schematic of our network architecture.

\PAR{2D image encoding.} We use a 2D encoder to map image patches to a latent representation. Each latent vector encodes an image patch associated with a 2D location. Given a query image $I\in \mathbb{R}^{H \times W \time 3}$, we use two levels of encodings: Coarse 2D features $F_{m}^{c} \in \mathbb{R} ^ {N_c \times C_c}$, where $N_c = \frac{H}{8} \times \frac{W}{8}$, and fine 2D features $F_{m}^{f} \in \mathbb{R} ^ {N_f \times C_f}$, where $N_f = \frac{H}{2} \times \frac{W}{2}$. $C_c$ and $C_f$ denote the encoding dimensions.

\PAR{3DGS filtering.}
Since the number of Gaussians can be very large (typically 700K), it is impractical to match all of them with the 2D features. Therefore, we first filter Gaussians with low opacity value and then randomly select a subset of $\hat{N}_g$ Gaussians.

\PAR{3DGS encoding.} We aim to match the locations of 3D Gaussians with image patches by aligning their features. Since the information in a single Gaussian is not rich enough to enable accurate matching, we construct an encoding incorporating also information from nearby Gaussians. To this end, we use KPFCNN \cite{thomas2019kpconv}, an encoder based on Kernel Point Convolution (KPConv) that applies convolution on a sphere. Additionally, KPFCNN incorporates downsampling layers that operate on a grid to reduce the number of 3D points to $N_g$. The encoder receives the parameters of the Gaussians as input, including their opacity, radiance (represented by spherical harmonic coefficients), orientation, and scaling. The outputs are features $F_{s} \in \mathbb{R}^{N_g \times C_c}$ associated with 3D points $Q \in \mathbb{R}^{N_g \times 3}$, such that each feature represents a 3D region of Gaussians. 

\PAR{3D-2D features alignment.}
Next, given the scene and image features, we apply a series of interleaved self- and cross-attention layers. The self-attention layers operate independently on the 3D and 2D features, allowing each domain to refine its representations and capture local and global dependencies. The cross-attention layers enable interaction between the 3D and 2D features, aligning the two domains by focusing on shared information. This interleaved process allows for an iterative update of feature representations, facilitating precise 3D-2D correspondences.

\PAR{Coarse-level matching.} 
Next, we compute pairwise cosine similarities between the coarse image features \( F_m^c \) and the 3DGS features \( F_s \), after we map them to a shared feature space, obtaining a matching score matrix \( \mathcal{S} \in \mathbb{R}^{N_{m} \times N_{s}} \). We then select matches whose scores exceed a threshold \( \theta_{c} \) while applying the mutual nearest neighbor (MNN) criterion to filter out potential outliers, yielding the final set of coarse correspondences:
\begin{equation*}
    \mathcal{M}_{c} = \{(i, j) \mid (i, j) \in \operatorname{MNN}(\mathcal{S}), \, \mathcal{S}(i, j) \geq \theta_{c} \}.
    \label{eq:coarse-matches}
\end{equation*}

\PAR{Coarse-to-Fine Module.} For each coarse match \( m_c = (i, j) \in \mathcal{M}_{c} \), we consider the fine-level feature \( F_m^f(i) \in \mathbb{R}^{w \times w \times D^f} \) corresponding to an image patch of size \( w \times w \) centered at the match location. We next apply a self-attention layer individually to each patch \( F_m^f(I) \) to enhance contextual information, while in parallel, we apply a linear layer to the coarse scene features to match the fine-level dimension \( D^f \).

Finally, we align the scene features with their corresponding fine image features. To that end, For each scene point, \( X_j \), we generate a heatmap expressing the probability that it aligns with each pixel near the image location \( x_i \).

The refined match location is then determined by taking an expectation over this probability distribution, resulting in the final fine-grained matches, denoted by \( \mathcal{M}_f \).

\subsection{Training}
For training, we use query images with known camera poses extracted from a Structure-from-Motion pipeline. We also use the output of this SfM pipeline to train our Gaussian Splatting scene representations. 
We train our model in three different pipelines: In \textit{Single-scene training}, we train a model on the Gaussians and images of a single scene and evaluate the model on test images from the same scene (in-scene generalization). In \textit{Multi-scene training}, we train a model on the Gaussians and images of a collection of scenes and evaluate the model on test images from the \textit{same} set of scenes. Lastly, for \textit{Cross-scene generalization}, we train a model on the Gaussians and images of a collection of scenes and evaluate the model on test images from \textit{novel} unseen scenes.

\PAR{Ground-truth matches.} 
We obtain ground-truth matches for training by projecting the 3DGS points $X_s$ onto the query image using the ground-truth query camera pose and use their projected locations as their fine-level matches.
For the coarse ground-truth matches, we compute a binary coarse association matrix $M_{gt}$ as follows. For each 3D point $X_j$, we use its projected location to assign it to the $i^{th}$ $8 \times 8$ patch to which it was matched, letting $M_{gt}(i, j) = 1$ if $X_j$ projects inside the 2D patch $i$. Notice that a 3D point can match at most one image patch, yet a single image patch can match multiple 3D points.

\PAR{Losses.}
The final loss consists of the losses for the coarse-level
and the fine-level: $L = L_c + L_f$:

\PAR{Coarse Loss.}
To guide the coarse matching, we apply the log loss from~\cite{sun2021loftr}. This will work to increase the dual-softmax probability at the ground-truth matching locations in $M_{gt}$. This loss is defined as
\begin{equation}\label{eq:coarse_matching_loss}
L_c = - \frac{1}{|\mathcal{M}_{gt}|} \sum_{(i, j) \in \mathcal{M}_{gt}} \log (S(i, j)).
\end{equation}
where \( |\mathcal{M}_{gt}| \) is the total number of ground-truth coarse matches, and \( S(i,j) \) denotes the matching score between the image patch \( i \) and the 3D point \( j \).

\PAR{Fine Loss.}
Let \( X_j \) and \( x_j \) respectively denote a 3D point and its ground-truth match, obtained by projecting \( X_j \) using the ground-truth camera pose. We set the fine matching loss to minimize the pixel distance between the predicted location, \( \tilde{x}_j \), and the ground-truth location, \( x_j \), and, following \cite{wang2020learning, sun2021loftr}, weigh this distance by the total variance \( \sigma^{2}(j) \) of the corresponding heatmap to penalize more heavily deviations in matches that are more certain. The loss is given by
\begin{equation}\label{eq:fine_matching_loss}
L_f = \frac{1}{|M_{f}|} \sum_{(i, j)\in M_f} \frac{1}{\sigma^2(j)} ||\tilde{x}_j - x_j||_2.
\end{equation}
where \( |\mathcal{M}_{f}| \) denotes the total number of predicted fine matches.

\subsection{Pose refinement}
\label{subsec:pose_refinment}

We follow \gscpr~\cite{liu2025gscpr} for pose refinement. Starting from an estimated pose of the query image, we use the 3DGS representation to render both an image and a depth map. We next use \mastr~\cite{leroy2024grounding} to establish dense 2D-2D correspondences between the query image and the rendered image. The matched points in the rendered image are lifted to 3D using the rendered depth, the estimated pose, and camera intrinsics, creating 3D-2D correspondences for the query image. Finally, we apply \pnp with RANSAC~\cite{fischler1981random} to solve for the refined pose.

 % \FloatBarrier

\begin{table*}[t!]

  \centering
  \setlength{\tabcolsep}{3pt}

\resizebox{1.0\textwidth}{!} {  
  \begin{tabular}{cllccccccccc}
    \toprule
    \multirow{2}{*} & \multirow{2}{*}{Method} & Scene &\multicolumn{9}{c}{7-Scenes - SfM Poses - Indoor} \\
    % \cmidrule(r){3-11}
    & & Repres. &  Chess & Fire & Heads & Office & Pump. & Kitchen & Stairs &  Avg.Med$\downarrow$ & Avg.Recall$\uparrow$.\\
    
\midrule
    \multirow{5}{*}{\rotatebox{90}{End-to-End}}
    & MS-Trans.~\cite{shavit2021learning} & APR Net.    
    & 11/6.4 & 23/11.5 & 13/13 & 18/8.1 & 17/8.4 & 16/8.9 & 29/10.3 & 18.1/9.5 & - \\
    & DFNet~\cite{chen2022dfnet} & APR Net.
    & 3/1.1 & 6/2.3 & 4/2.3 & 6/1.5 & 7/1.9 & 7/1.7 & 12/2.6  & 6.4/1.9 & -\\
    & NeFeS~\cite{chen2023refinement} & APR+NeRF
    & 2/0.8 & 2/0.8 & 2/1.4 & 2/0.6 & 2/0.6 & 2/0.6 & 5/1.3 & 2.4/0.9 & - \\
    
    &  DSAC*~\cite{brachmann2021visual} & SCR Net. 
    & 0.5/0.2 & 0.8/0.3	& 0.5/0.3 & 1.2/0.3	& 1.2/0.3 & 0.7/0.2 & 2.7/0.8 & 1.1/0.3 & \bf{97.8} \\
    & ACE~\cite{brachmann2023accelerated} & SCR Net. 
    &0.7/0.5 & 0.6/0.9 & 0.5/ 0.5 & 1.2/0.5	& 1.1/0.2 & 0.9/0.5	& 2.8/1.0 & 1.1/0.6 & 97.1 \\
    % \cmidrule(r){3-11}
    \midrule
    \multirow{3}{*}{\rotatebox{90}{Hierarchical}}
    & DVLAD+R2D2\cite{torii201524, revaud2019r2d2} & 3D+RGB 
    &  0.4/\bf{0.1} & \bf{0.5/0.2} & \bf{0.4/0.2} & \bf{0.7/0.2} & \bf{0.6/0.1} & \bf{0.4/0.1} & \bf{2.4/0.7} & \bf{0.8/0.2} & 95.7 \\
    & HLoc\cite{sarlin2019coarse} & 3D+RGB    
    & 0.8/\bf{0.1} & 0.9/\bf{0.2}	& 0.6/0.3 & 1.2/\bf{0.2} & 1.4/0.2 & 1.1/\textbf{0.1}	& 2.9/0.8 & 1.3/0.3 & 95.7  \\
    % \cmidrule(r){3-11}
   & NeRFLoc~\cite{liu2023nerf} & NeRF+RGBD    
    &  2/1.1 & 2/1.1 & 1/1.9 & 2/1.1 & 3/1.3 & 3/1.5 & 3/1.3 & 2.3/1.3 & - \\
    & \nerfmatch \cite{zhou2024nerfect} & NeRF+RGB    
    &  0.9/0.3 & 1.1/0.4 & 1.4/1.0 & 3.0/0.8 & 2.2/0.6 & 1.0/0.3 & 9.0/1.5 & 2.7/0.7 & 78.2 \\
    
    \midrule 
    \multirow{3}{*}{\rotatebox{90}{GS-based}}
   & 6DGS \cite{bortolon20246dgs} & 3DGS
    &  26.8/28.7 & 33.3/36.8 & 17.3/33.7 & 37.6/31.0 & 22.1/28.0 & 42.5/35.7 & 47.5/31.7 & 32.4/32.2 & - \\
    & GSplatLoc \cite{sidorov2024gsplatloc}& 3DGS
    &  0.43/0.16 & 1.03/0.32 & 1.06/0.62 & 1.85/0.4 & 1.8/0.35 & 2.71/0.55 & 8.83/2.34 & 2.53/0.68 & -\\
    & \textbf{\method (Ours)} & 3DGS
    &  \textbf{\underline{0.39}}/\underline{0.13} & \underline{0.58}/\underline{0.24} & \underline{0.54}/\underline{0.34} & \underline{1.0}/\underline{0.26} & \underline{0.90}/\underline{0.21} & \underline{0.73}/\underline{0.18} & \underline{4.7}/\underline{0.96} & \underline{1.3}/\underline{0.33} & 88.0 \\
    
    \bottomrule
  \end{tabular}
  }
\caption{\textbf{Indoor Localization on 7-Scenes~\cite{glocker2013real, shotton2013scene}.} We report per-scene median position errors (in cm) and rotation errors (in degrees), along with their averages across scenes and the mean localization recall. The best result in each column is highlighted in \textbf{bold}, and the best result for the \gs representation is \underline{underlined}.}

    \label{tab:loc_7sc}
\end{table*}

\begin{table*}[!t]

\centering
%\setlength{\tabcolsep}{1pt}
%\scriptsize
\footnotesize

%\resizebox{0.49\textwidth}{!} {  
\begin{tabular}{cllccccc}
    \toprule
    \multicolumn{2}{c}{\multirow{2}{*}{Method}} & Scene  & \multicolumn{5}{c}{Cambridge Landmarks - Outdoor}  \\  
    \cmidrule{4-8}   
    & & Repres.
    & Kings & Hospital & Shop & StMary & Avg.Med$\downarrow$ \\
    \midrule
    \multirow{7}{*}{\rotatebox{90}{End-to-End}}
    &MS-Trans.~\cite{shavit2021learning} & APR Net.
    & 83/1.5 & 181/2.4 & 86/3.1 &  162/4 & 128/2.8 \\
    &DFNet~\cite{chen2022dfnet} & APR Net.
    & 73/2.4 & 200 /3 & 67/2.2 & 137/4 & 119.3/2.9 \\
    &LENS~\cite{moreau2022lens} & APR Net.
    & 33/0.5 & 44/0.9 & 27/1.6 & 53/1.6	& 39.3/1.2	\\
    &NeFeS~\cite{chen2023refinement} & APR+NeRF
    & 37/0.6 & 55/0.9 & 14/0.5 & 32/1 & 34.5/0.8 \\
    &DSAC*~\cite{brachmann2021visual}  & SCR Net.
    & 15/0.3 & 21/0.4 & 5/0.3 & 13/0.4 & 13.5/0.4 \\
    &HACNet~\cite{li2020hierarchical} & SCR Net.
    & 18/0.3 & 19/\bf{0.3} & 6/0.3 & 9/0.3 & 13/0.3 \\    
    &ACE~\cite{brachmann2023accelerated} & SCR Net. 
    & 28/0.4 & 31/0.6 & 5/0.3 & 18/0.6 & 20.5/0.5 \\
    \midrule                                        
    \multirow{11}{*}{\rotatebox{90}{Hierarchical}}    
    &SANet~\cite{yang2019sanet} & 3D+RGB     
    & 32/0.5 & 32/0.5 & 10/0.5 & 16/0.6 & 22.5/0.5  \\    
    &DSM~\cite{tang2021learning} & SCR Net.          
    & 19/0.4 & 24/0.4 & 7/0.4 & 12/0.4 & 15.5/0.4  \\
    &NeuMap~\cite{tang2023neumap} & SCode+RGB   
    & 14/\bf{0.2} & 19/0.4	& 6/0.3 & 17/0.5 & 14/0.3 \\
    &InLoc\cite{taira2018inloc} & 3D+RGB    
    & 46/0.8 & 48/1.0 & 11/0.5 & 18/0.6 & 30.8/0.7 \\    
    &HLoc\cite{sarlin2019coarse} & 3D+RGB
    &12/\bf{0.2} & \bf{15/0.3} & \bf{4/0.2} & \bf{7/0.2} & \bf{9.5}/\bf{0.2} \\
    &PixLoc\cite{sarlin2021back} & 3D+RGB
    & 14/\bf{0.2} & 16/\bf{0.3} & 5/\bf{0.2} &10/0.3 & 11/0.3 \\
    &CrossFire~\cite{moreau2023crossfire} & NeRF+RGB
    & 47/0.7 & 43/0.7 & 20/1.2 & 39/1.4 & 37.3/1 \\
    &NeRFLoc~\cite{liu2023nerf} & NeRF+RGBD
    & \bf{11/0.2} & 18/0.4 & \bf{4/0.2} & \bf{7/0.2} & 10/0.3 \\

    &\nerfmatch \cite{zhou2024nerfect} & NeRF+RGB    
    & 13.0/\bf{0.2}	& 19.4/0.4 & 8.5/0.4 & 7.9/0.3 & 12.2/0.3 \\

        \midrule  
    & GSplatLoc \cite{sidorov2024gsplatloc} & 3DGS  &  27/0.46 & \underline{20}/0.71 & \underline{5}/0.36 & 16/0.61 & \underline{17}/0.53 \\
    & \textbf{\method (Ours)} & 3DGS & \underline{23}/\underline{0.3} & 22/\underline{0.42} & 8/\underline{0.29} & \underline{14}/\underline{0.45} & \underline{17}/\underline{0.36} \\
    \bottomrule
  \end{tabular}

%}
\caption{\textbf{Outdoor Localization on Cambridge Landmarks~\cite{kendall2015posenet}.} We report per-scene median position errors (in cm) and rotation errors (in degrees), along with the averages across scenes. The best result in each column is highlighted in \textbf{bold}, while the best result for the \gs representation is \underline{underlined}.} 

\label{tab:loc_camb}
\end{table*}

\section{Experiments}
\label{sec:experiments}

\subsection{Experimental setup}
\label{subsec:setup}

\PAR{Datasets.} We conduct \emph{single-scene} training on two well-established localization datasets. The \textbf{7-Scenes} dataset \cite{glocker2013real, shotton2013scene} consists of RGB-D images captured across seven unique indoor scenes (volumes ranging from $1m^3$ to $18m^3$) that are challenging due to the presence of texture-less surfaces, motion blur, and occlusions. We follow the original train/test splits and use more accurate SfM pose annotations, as recommended by \cite{brachmann2023accelerated, brachmann2021limits, chen2023refinement}, for both our method and all baselines.

The \textbf{Cambridge Landmarks} dataset \cite{kendall2015posenet} contains handheld smartphone images from outdoor scenes, each characterized by significant exposure variations that complicate large-scale localization. Consistent with previous works \cite{trivigno2024unreasonable, sidorov2024gsplatloc}, we evaluate on King's College, Old Hospital, Shop facade, and St Mary's church (spanning \(875-5600m^{2}\)), following the original splits.

For \emph{cross-scene generalization}, we use \textbf{\scannetp} \cite{yeshwanthliu2023scannetpp}, a large-scale indoor dataset that couples high-quality (sub-millimeter laser scans) and commodity-level (registered images) geometry and color captures. Unlike setups that merely hold out a portion of training images along the same camera trajectory, \scannetp\ provides novel test views per scene, leading to more realistic and challenging conditions for cross-scene visual localization. The dataset encompasses large and varied indoor environments with numerous glossy and reflective surfaces, making accurate localization particularly difficult. We randomly sample 130 scenes for training, 7 for validation, and 15 for testing.

\PAR{Evaluation metrics.} In line with previous works, we report median pose errors, specifically translation error in centimeters and rotation error in degrees. On the 7-Scenes dataset, we also report localization recall, which measures the percentage of query images localized with pose errors below specified thresholds—namely, \(5\text{cm}\) for the translation and \(5^\circ\) for the rotation.

\PAR{Baselines.}  We compared our method to MS-Trans.~\cite{shavit2021learning}, DFNet~\cite{chen2022dfnet}, LENS~\cite{moreau2022lens}, NeFeS~\cite{chen2023refinement}, DSAC*~\cite{brachmann2021visual}, HACNet~\cite{li2020hierarchical}, ACE~\cite{brachmann2023accelerated}, SANet~\cite{yang2019sanet}, DSM~\cite{tang2021learning}, NeuMap~\cite{tang2023neumap}, InLoc\cite{taira2018inloc}, HLoc\cite{sarlin2019coarse}, PixLoc\cite{sarlin2021back}, CrossFire~\cite{moreau2023crossfire}, NeRFLoc~\cite{liu2023nerf}, \nerfmatch \cite{zhou2024nerfect}, GSplatLoc \cite{sidorov2024gsplatloc}, DVLAD+R2D2\cite{torii201524, revaud2019r2d2} and 6DGS \cite{bortolon20246dgs}. Following \cite{zhou2024nerfect}, we categorize the methods into three groups: end-to-end methods, which include APR and SCR methods; hierarchical methods, where an initialized camera pose is estimated using an image retrieval step; and 3D\gs-based methods. Additionally, we include the scene representation used for localization during testing. For experiments on Cambridge and 7-scenes, all the evaluations of the baseline methods were taken from previous works except for 6DGS \cite{bortolon20246dgs}, which we trained and evaluated using the official implementation. Since \scannetp is a relatively new dataset, there are no publicly available baseline evaluations. Therefore, we compared our cross-scene generalization capabilities to both GSPlatLoc and our method in a single-scene training pipeline.

\PAR{Implementation details.} 
We use the first two blocks of ConvFormer~\cite{yu2023metaformer} as the image backbone, initialized with ImageNet-1K~\cite{russakovsky2015imagenet} pre-trained weights\footnote{Pre-trained weights can be found here: \url{huggingface.co/timm/convformer_b36.sail_in1k_384}}. The feature dimensions are set to \( D^c = 512 \) for the coarse matching and \( D^f = 128 \) for the fine matching. For fine matching, a local window size of \( w = 5 \) is used for image feature cropping. $\theta_{c}=0.3$ the score threshold for the coarse matches.  Query images are resized to \( 480 \times 480 \) in all experiments.
We filter Gaussians with opacity values lower than 0.9 and then uniformly subsample 100K Gaussians. For the 3D backbone, we use a 3-stage KPFCNN \cite{thomas2019kpconv} with output channels of \{128, 256, 512\} for each stage. We use the output of the final layer for encoding.
The 3D-2D alignment module is composed of four interleaved self- and cross-attention layers. Both KPFCN and 3D-2D alignment modules are initialized with random weights.
For pose refinement, we re-implemented \gscpr, as the code is not yet publicly available. Unlike their approach, we use the original Gaussian splatting model \cite{kerbl3Dgaussians} and omit the exposure-adaptive transformation. At inference, we apply three refinement iterations for each query image.

We train our models using the Adam optimizer~\cite{kingma2014adam} with a learning rate of \( \text{lr} = 0.0001 \) and batch size \( \text{bs} = 1 \) for 100 epochs. Our models are trained on a single Nvidia A40 GPU (48 GB). 
On a Quadro RTX~6000 and \scannetp images at $1752{\times}1168$, our average per-query time is $2.82$\,s ($1.03$\,s inference + $1.79$\,s refinement). The current implementation is not fully optimized and can be further accelerated.
Further implementation details of the 3D\gs are provided in the supplementary material.

\begin{figure*}[!t]
    \centering
    \includegraphics[width=1.0\linewidth, trim={0cm 0cm 0cm 0cm}]{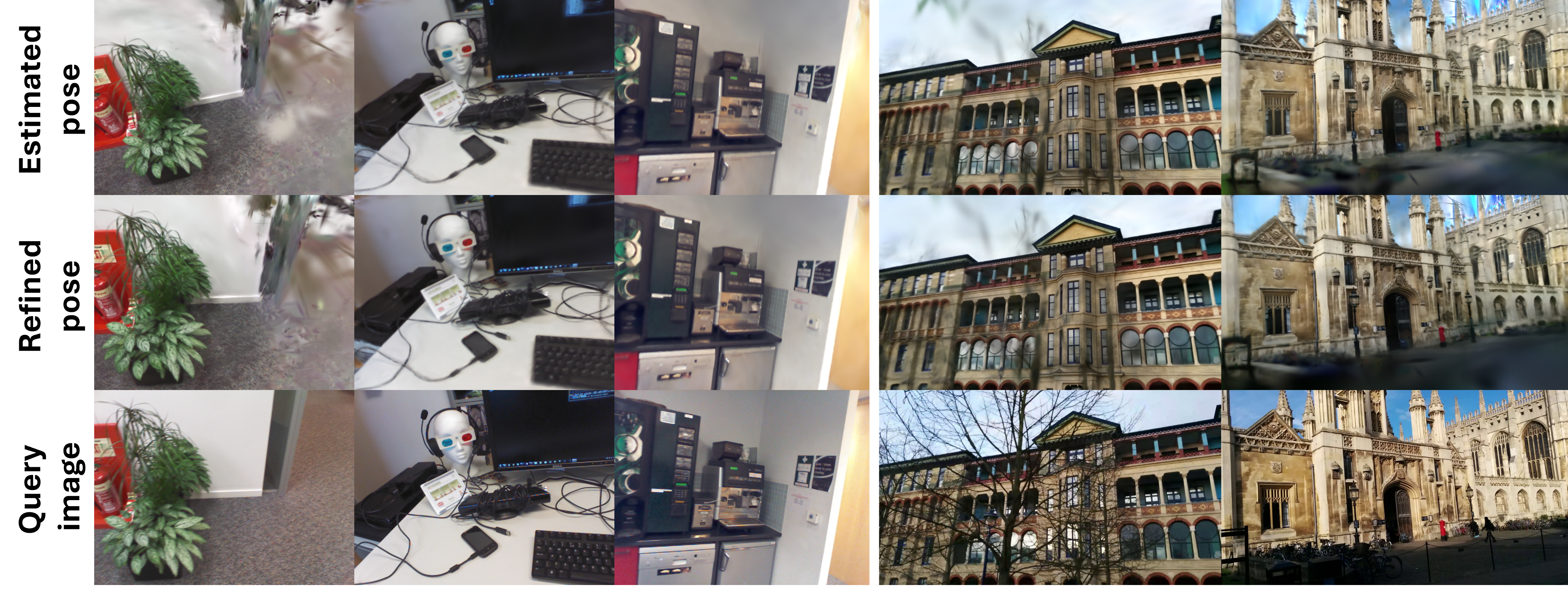}
    \caption{\textbf{\method Qualitative Results.} Visualization of our model’s pose estimation on indoor scenes (three left columns) and outdoor scenes (two right columns). Each row, from top to bottom, displays rendered images from the model's estimated pose, rendered images from the refined pose, and the corresponding query images.}
    \label{fig:visual}
\end{figure*}

\begin{table*}[t!]
    \centering
    \setlength{\tabcolsep}{4pt}

    \begin{tabular}{lccccccccc}
    \toprule
    Method &  Chess & Fire & Heads & Office & Pump. & Kitchen & Stairs &  Avg.Med$\downarrow$ & Avg.Recall$\uparrow$.\\
    
    \midrule
    Single scene  &  \textbf{0.39}/\textbf{0.13} & \textbf{0.58}/\textbf{0.24} & \textbf{0.54}/\textbf{0.34} & \textbf{1.0}/\textbf{0.26} & \textbf{0.90}/\textbf{0.21} & \textbf{0.73}/\textbf{0.18} & 4.7/0.96 & \textbf{1.3}/\textbf{0.33} & \textbf{88.0} \\
    Multi. scene  &  0.59/0.15 &0.73/0.28 & 0.57/0.37 & 1.5/0.36& 1.0/\textbf{0.21}& 1.1/0.26 &\textbf{4.5}/\textbf{0.9} & 1.4/0.36 & 84.4\\
    \bottomrule
    \end{tabular}
    \caption{\textbf{Multi-scene Model on 7-Scenes}~\cite{glocker2013real, shotton2013scene}. We evaluate our multi-scene model against a single-scene model, reporting per-scene median position errors (cm) and rotation errors (degrees), their averages across scenes, and the mean localization recall. The best result in each column is shown in \textbf{bold}.}
    
    \label{tab:multiple_scenes}
\end{table*}

\PAR{Hyper-parameter search.}
We determined the values of hyper-parameters by running our models on the validation split from the 7-scenes datasets. These hyper-parameters include the learning rate, batch size, number of interleaved self- and cross-attention layers in the 3D-2D alignment module, and the number of layers used by the 3D encoder.

\subsection{Single-scene training}
\label{subsec:single_scene_results}
\PAR{Results on the 7-Scenes Dataset.}
As shown in Table~\ref{tab:loc_7sc}, our method outperforms other 3DGS-based approaches and achieves higher rotation and translation accuracy than APR and NeRF-based methods. It also performs on par with SCR methods, though it is slightly outperformed by the structure-based DVLAD+R2D2 \cite{revaud2019r2d2,torii201524}.

\PAR{Results on the Cambridge Landmarks  Dataset.}  As shown in Table~\ref{tab:loc_camb}, our approach surpasses GSplatLoc, a concurrent 3DGS-based method, in rotation accuracy while achieving comparable translation accuracy. It further outperforms APR methods by a significant margin in both rotation and translation accuracy. However, it trails behind the state-of-the-art method HLoc, particularly in translation accuracy. We believe this is due to the relatively low rendering quality of 3DGS on the Cambridge Landmarks dataset (see supplementary material). Improved 3DGS representations optimized for such challenging scenes may help bridge this gap.

\PAR{Qualitative Results.}
In Figure~\ref{fig:visual}, we provide qualitative results demonstrating the accuracy of the pose estimation achieved with our method both before and after the refinement stage. Each example includes the ground-truth (GT) query image and rendered images from the initially estimated pose and the final refined pose.

The initial pose estimates align reasonably well with the GT, though some misalignments remain. After refinement, the final pose rendering shows improved alignment, with details that closely match the GT image. This refinement step corrects subtle orientation and translation discrepancies. 

\subsection{Multi-scene training}
\label{subsec:multi_scene_results}
A multi-scene model can encode information from multiple scenes in a single model while retaining accuracy similar to single-scene training. In addition, a multi-scene model allows for lower resource usage at inference. We tested the ability of our model to encode several scenes simultaneously. To that end, we trained our model on training images from all the scenes in the 7-scenes dataset and tested it on test query images from the same scenes. Since the size of the training data for this experiment is large, we train \method without the fine-level matching (coarse-level only) and compare it to single-scene models with fine-level matching. The results in Table~\ref{tab:multiple_scenes} indicate that, despite training across multiple scenes at coarse-level, we observe only a slight reduction in prediction accuracy and even an improvement in the Stairs scene.

\begin{table}[t!]
\Large
\begin{adjustbox}{max width=\columnwidth} 
\begin{tabular}{lcc|c}
\toprule
\multirow{2}{*}{Scene/Method} & \multicolumn{2}{c}{Single-scene} & Cross-scene \\
\cmidrule{2-4}
 & GSplatLoc \cite{sidorov2024gsplatloc} & \multicolumn{1}{c}{GSVisLoc (Ours)} & GSVisLoc (Ours) \\
\midrule
c0da8f4a4d & 154.75/18.38 	& \textbf{0.31}/\textbf{0.08} & \underline{0.53}/\underline{0.14} \\
dfa70fb232 & 61.96/6.59 	& \textbf{0.71}/\underline{0.15} & \underline{0.74}/\textbf{0.14} \\
a16208dde3 & 6.97/1.62 & \underline{0.58}/\underline{0.15} & \textbf{0.57}/\textbf{0.14} \\
7f97f24691 & 12.08/1.54 & \underline{0.57}/\underline{0.09} & \textbf{0.41}/\textbf{0.07} \\
c8eeef6427 & 3.54/0.44 & \underline{0.75}/\textbf{0.16} & \textbf{0.59}/\underline{0.18} \\
ba89245cfd & 2.8/0.37 & \underline{0.58}/\underline{0.08} & \textbf{0.5}/\textbf{0.06} \\
578511c8a9 & 6.31/0.41 & \underline{0.41}/\textbf{0.07} & \textbf{0.38}/\textbf{0.07} \\
19cfd590f4 & 2.1/0.55 & \textbf{0.55}/\textbf{0.11} & \underline{0.59}/\textbf{0.11} \\
e3ad7115db & 1.54/0.43 & \textbf{0.58}/\textbf{0.1} & \underline{0.71}/\underline{0.11} \\
de5881aa12 & 7.82/1.6 & \underline{0.71}/\underline{0.23} & \textbf{0.51}/\textbf{0.16} \\
bf07750a0b & 1.74/0.42 & \underline{0.23}/\textbf{0.1} & \textbf{0.21}/\underline{0.14} \\
f9397af4cb & 2.81/0.4 & \textbf{0.27}/\underline{0.14} & \underline{0.29}/\textbf{0.11} \\
bde1e479ad & 13.11/1.1 & \textbf{0.59}/\underline{0.1} & \textbf{0.59}/\textbf{0.09} \\
7104910700 & \underline{14.28}/\underline{2.21} & \textbf{1.43}/\textbf{0.26} & 21.65/37.17 \\
a492fe77aa & 139.22/15.8 & \underline{0.9}/\underline{0.23} & \textbf{0.85}/\textbf{0.18} \\
\bottomrule
Avg.Med$\downarrow$ & 28.74/3.46 & \textbf{0.61}/\textbf{0.14} & \underline{1.94}/\underline{2.59} \\
\bottomrule
\end{tabular}
\end{adjustbox}
\caption{\textbf{Generalization on \scannetp \cite{yeshwanthliu2023scannetpp}.} We report the median position errors (in cm) and rotation errors (in degrees) for each scene, along with the overall averages across scenes. The best result in each row is shown in \textbf{bold}, and the second-best result is \underline{underlined}.}
\label{tab:generalization}
\end{table}
\subsection{Cross-scene generalization}
\label{subsec:generalization}
We next tested our model for cross-scene generalization. We trained the model on multiple scenes from the \scannetp dataset and tested it on novel, unseen scenes from the test split. We compared the cross-scene model to GSPlatLoc and to our method trained in the single-scene training pipeline. We note that these two baselines, unlike our cross-scene model, were trained specifically on the tested scenes. Due to shortage of resources, we trained this model without the fine-level matching, as in the multi-scene setup. As shown in Table ~\ref{tab:generalization}, our cross-scene model outperforms GSPlatLoc and achieves comparable results to our single-scene pipeline.

\begin{table}[t!]
\centering

    \begin{tabular}{lccc}
    \toprule
                                & Fire             & Heads            & Pumpkin \\
    \midrule
    No 3D encoder               &    -      &     -         & -\\
    No 2D encoder fine-tuning   & 3.7/1.4          & 4.6/2.7          & 4.1/0.70\\ 
    No 3D-2D alignment          & 2.7/\textbf{1.0} & 3.8/2.3          & 2.3/\textbf{0.39} \\
    No fine-level matching      & 6.8/2.5          & 4.9/2.8          & 3.4/0.56 \\
    \textbf{\method}                     & \textbf{2.6}/1.1 & \textbf{3.4/2.1} & \textbf{2.0/0.39} \\
    \bottomrule
    \end{tabular}
    \caption{\textbf{Ablation Studies.} We evaluate our model's performance by removing key components: the 3D encoder, fine-tuning of the 2D encoder, the 3D-2D alignment module, and find-level matching, compared to \method. We report per-scene median translation errors (in cm) and rotation errors (in degrees) on 7-Scenes without pose refinement. The best result in each column is highlighted in \textbf{bold}.}

\label{tab:ablation}
\end{table}

\subsection{Ablations}
\label{subsec:ablations}
We conducted ablation studies to demonstrate the importance of different components of our model. We tested our model without (a) a 3D encoder, (b) the fine-tuning of a 2D encoder, (c) a 3D-2D alignment module, and (d) fine-level matching. We report the results before pose refinement in Table~\ref{tab:ablation}. For (a), when the 3D encoder was omitted, we sampled \( N_g \) 3D points and used a single linear layer to align the feature dimensions between the Gaussians and the query image. In this setup, the model was unable to extract a sufficient number of matches for reliable camera pose estimation, highlighting the necessity of encoding a 3D region rather than relying on individual Gaussians. Training a model without either fine-level matching or fine-tuning the 2D encoder hurts the model's accuracy, demonstrating their importance. Surprisingly, removing the 3D-2D alignment module (i.e., using dual-softmax directly on the encoder outputs, without applying self and cross-attention layers) had minimal impact on the results in a single-scene training setup, indicating that the encoders effectively extract corresponding 3D and 2D features. However, in the cross-scene generalization setup, omitting the 3D-2D alignment module led to near-random pose estimates.

 \section{Conclusion}
\label{sec:conclusion}

In this work, we present GSVisLoc, a generalizable visual localization method for a 3D Gaussian Splatting (3DGS) scene representation, whose task is to estimate the camera position and orientation in a 3D environment given a query image. Our method, which uses deep learning to establish 3D-2D correspondences enables effective localization with 3DGS as the sole scene representation, eliminating the need for reference or training images during inference.
Our method achieves competitive localization accuracy on standard benchmarks that include images from both indoor and outdoor scenes. The coarse-to-fine matching strategy, coupled with a 3DGS-based pose refinement step, allows for precise pose estimation. Most importantly, our method is trainable, enabling cross-scene generalization with almost no loss of accuracy. By that, our work opens avenues for visual localization for 3DGS scene representation, without reliance on additional reference data.
Still, our results highlight the importance of improving 3DGS scene representation on outdoor scenes.

 \section*{Acknowledgments}
\label{sec:acknowledgments}

Research was partially supported by the Israeli Council for Higher Education (CHE) via the Weizmann
Data Science Research Center, by the MBZUAI-WIS Joint Program for Artificial Intelligence
Research and by research grants from the Estates of Tully and Michele Plesser and the Anita James
Rosen and Harry Schutzman Foundations
 % \clearpage

  {
    \small
    \bibliographystyle{ieeenat_fullname}
    \bibliography{main}
}

\clearpage
\setcounter{page}{1}
\maketitlesupplementary

\begin{table}[t!]
    \caption{\textbf{Gaussian splatting PSNR scores.} We provide the PSNR scores for our trained \gs models across each scene in the Cambridge Landmarks~\cite{kendall2015posenet} and 7-Scenes~\cite{glocker2013real, shotton2013scene} datasets.}
    \resizebox{0.48\textwidth}{!} {  
        \begin{tabular}{ccccc}
        \multicolumn{5}{l}{\normalsize Cambridge Landmarks - Outdoor} \\
        \toprule
        Kings & Hospital & Shop & StMary  & Average \\
        \midrule    
        20.8& 16.9 & 22.0 & 21.7 & 20.4   \\
        \bottomrule
        \end{tabular}
    }

    \bigskip\bigskip
    \resizebox{0.48\textwidth}{!} {  
        \begin{tabular}{cccccccc}
        \multicolumn{8}{l}{\Large 7-Scenes - Indoor} \\
        \toprule
        Chess & Fire & Heads & Office & Pump. & Kitchen & Stairs & Average\\
        \midrule    
        28.9&28.8  &30.7 & 28.9& 30.0 &24.4 & 30.4 &  28.9\\
        \bottomrule
        \end{tabular}
    }
    \label{tab:gs_psnr}
\end{table}

\begin{figure*}[h]
    \centering
    % \begin{tabular}{m{2cm} c}  % Adjust the width (2cm) as needed
    \begin{tabular}{lcccc}
        \rotatebox{90}{\textbf{52599ae063}} &  % Rotated vertical text
        \includegraphics[width=0.22\textwidth]{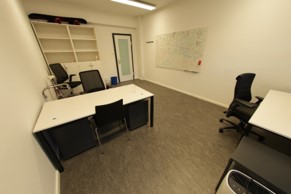} &
        \includegraphics[width=0.22\textwidth]{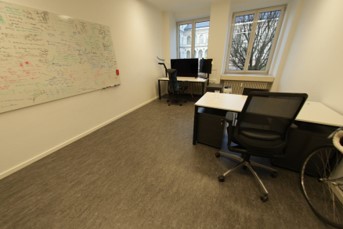} &
        \includegraphics[width=0.22\textwidth]{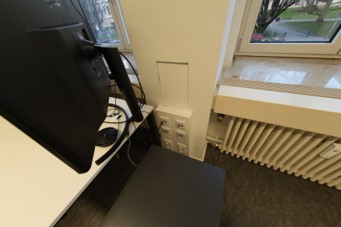} &
        \includegraphics[width=0.22\textwidth]{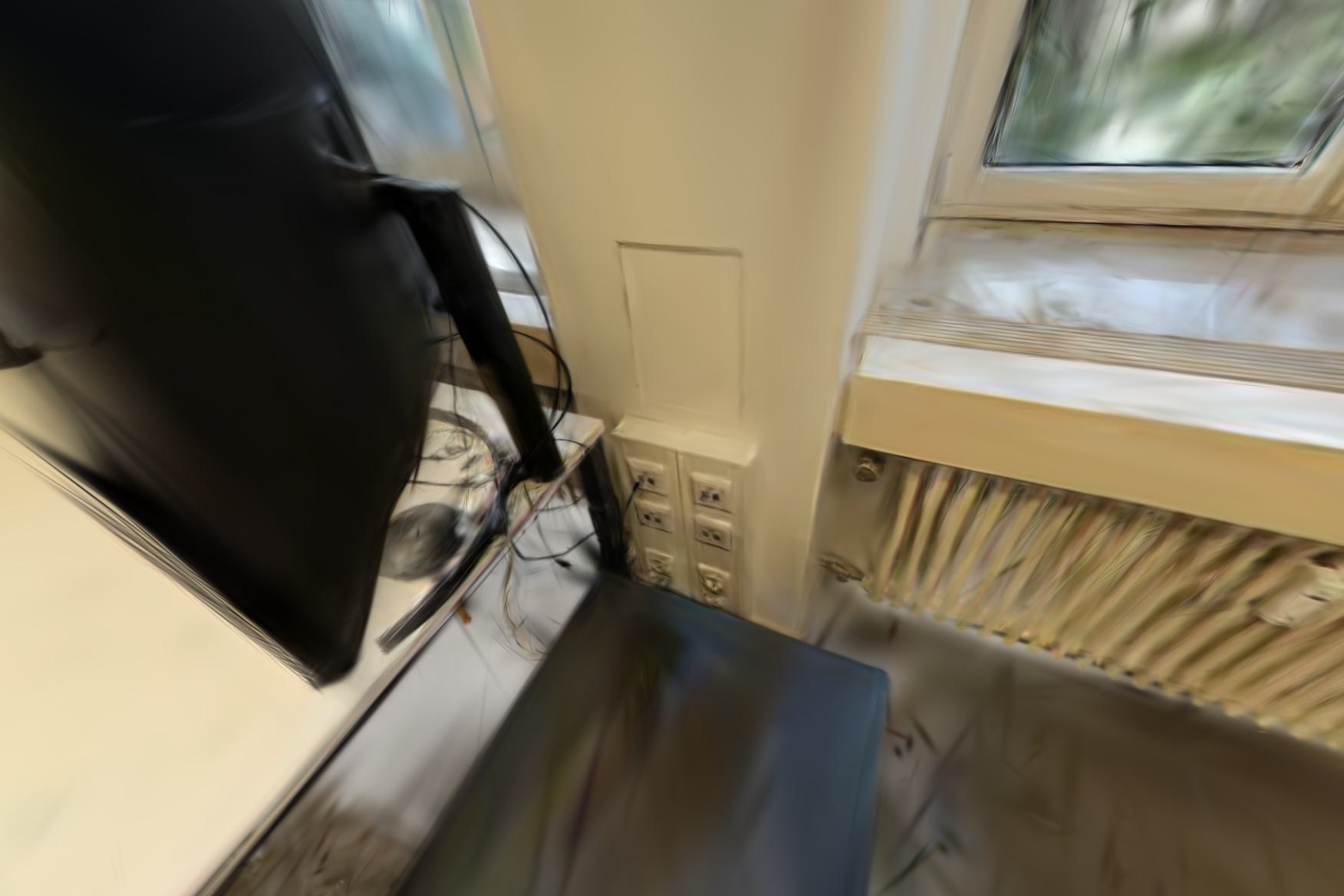} \\ 

        \rotatebox{90}{\textbf{b26e64c4b0}} &  % Rotated vertical text
        \includegraphics[width=0.22\textwidth]{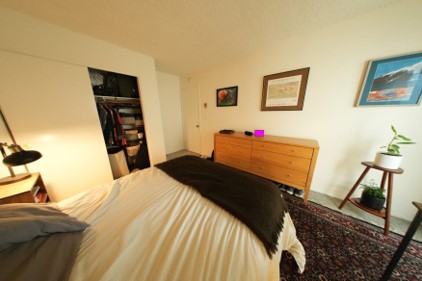} &
        \includegraphics[width=0.22\textwidth]{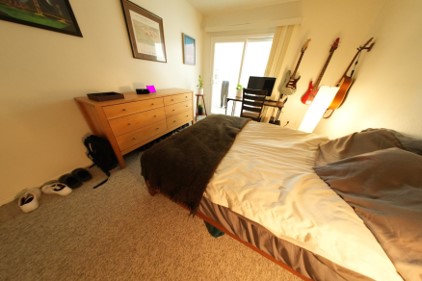} &
        \includegraphics[width=0.22\textwidth]{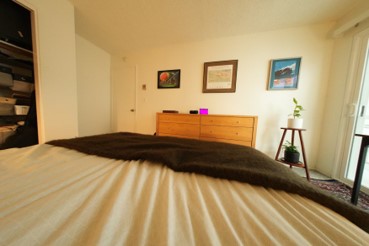} &
        \includegraphics[width=0.22\textwidth]{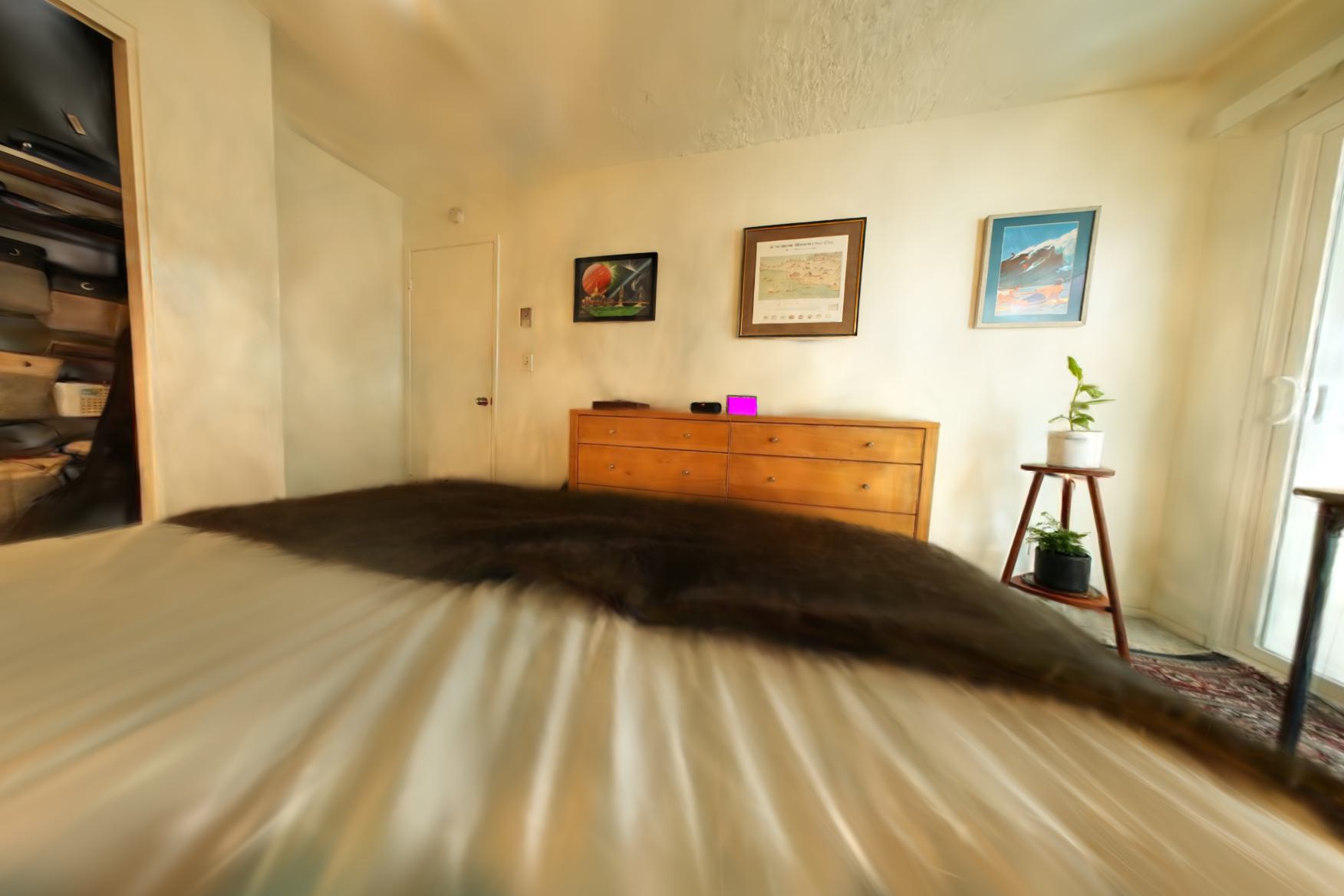} \\ 

        \rotatebox{90}{\textbf{1204e08f17}} &  % Rotated vertical text
        \includegraphics[width=0.22\textwidth]{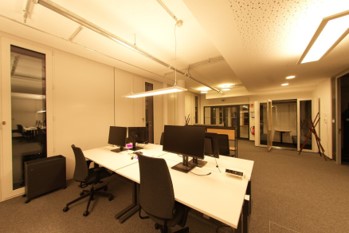} &
        \includegraphics[width=0.22\textwidth]{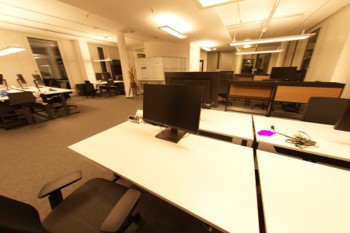} &
        \includegraphics[width=0.22\textwidth]{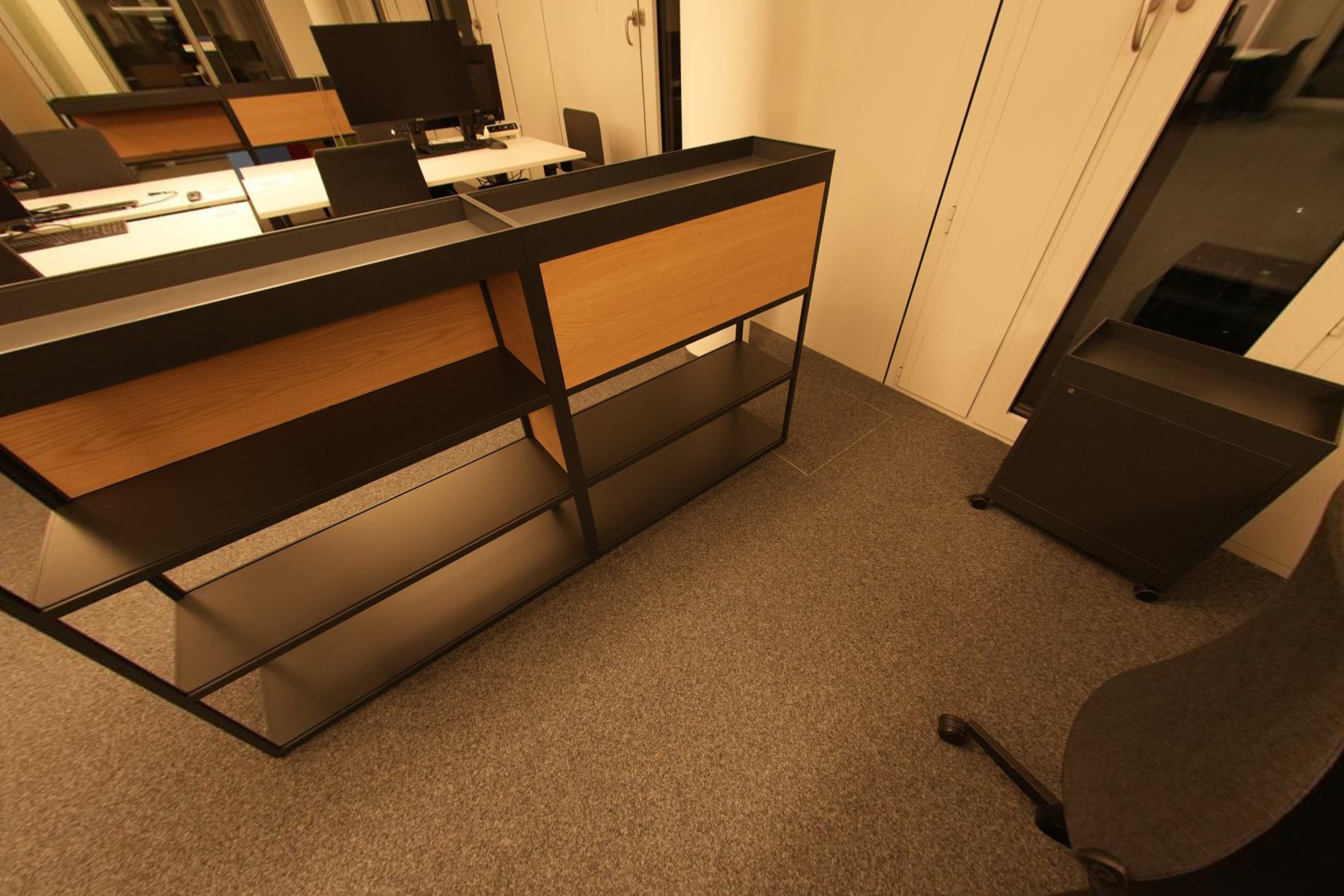} &
        \includegraphics[width=0.22\textwidth]{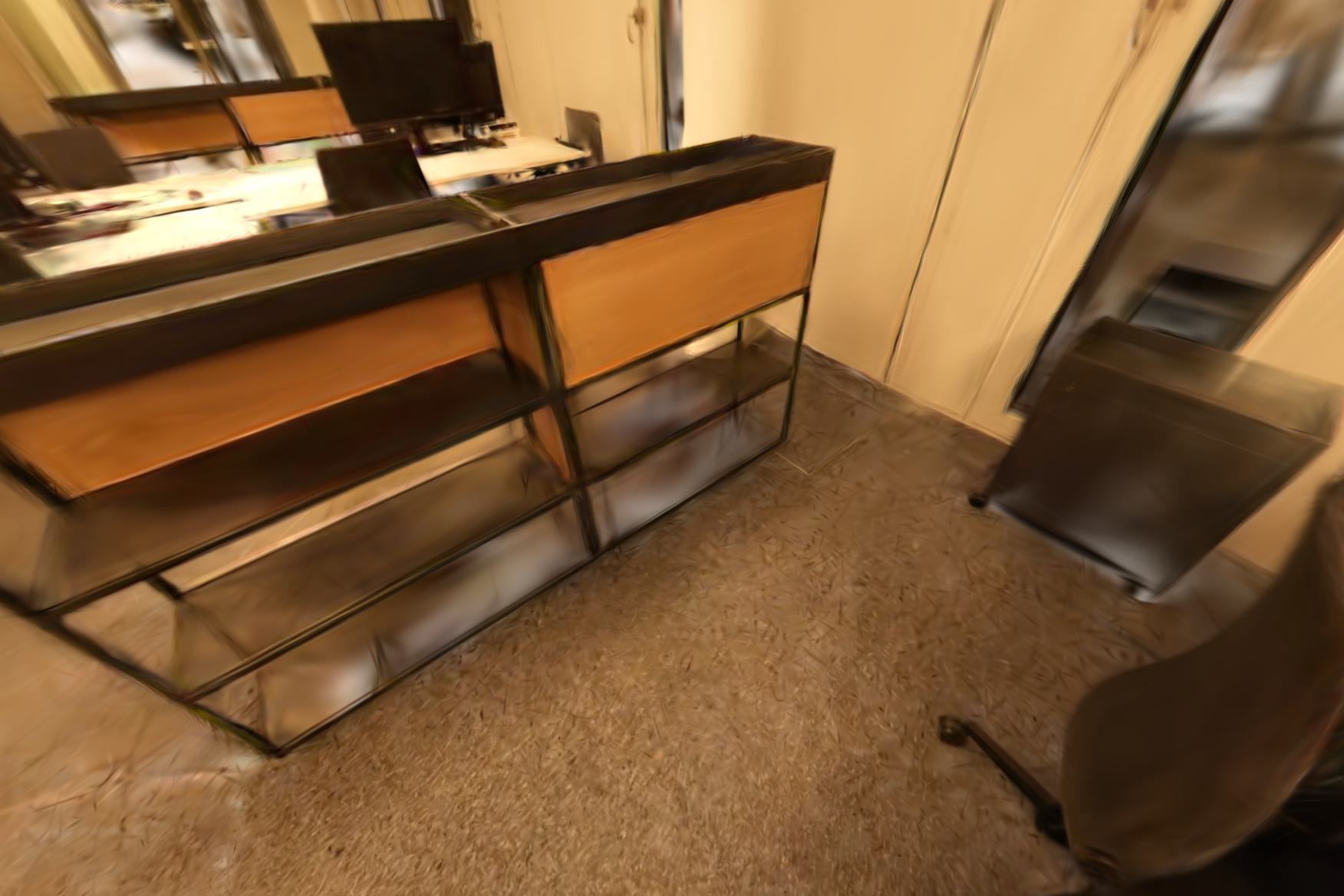} \\ 

        \rotatebox{90}{\textbf{480ddaadc0}} &  % Rotated vertical text
        \includegraphics[width=0.22\textwidth]{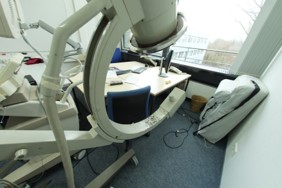} &
        \includegraphics[width=0.22\textwidth]{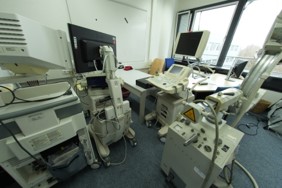} &
        \includegraphics[width=0.22\textwidth]{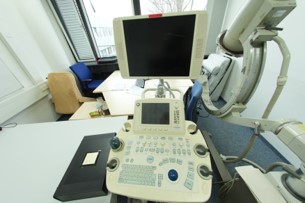} &
        \includegraphics[width=0.22\textwidth]{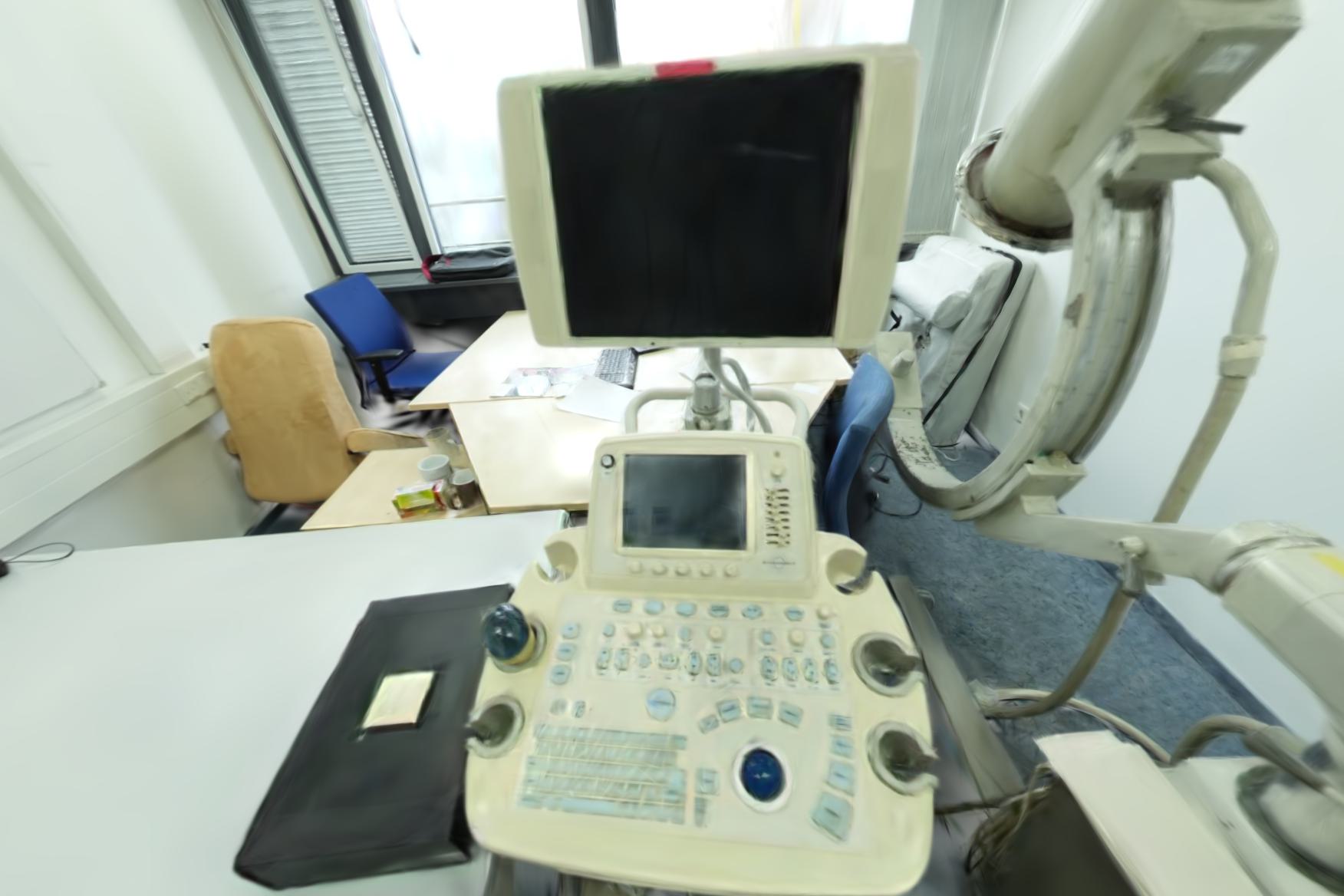} \\ 

        \rotatebox{90}{\textbf{ab11145646}} &  % Rotated vertical text
        \includegraphics[width=0.22\textwidth]{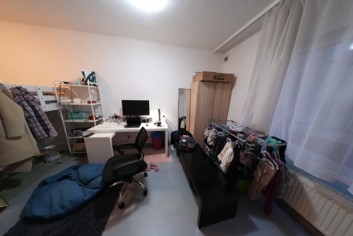} &
        \includegraphics[width=0.22\textwidth]{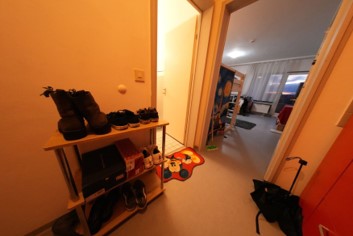} &
        \includegraphics[width=0.22\textwidth]{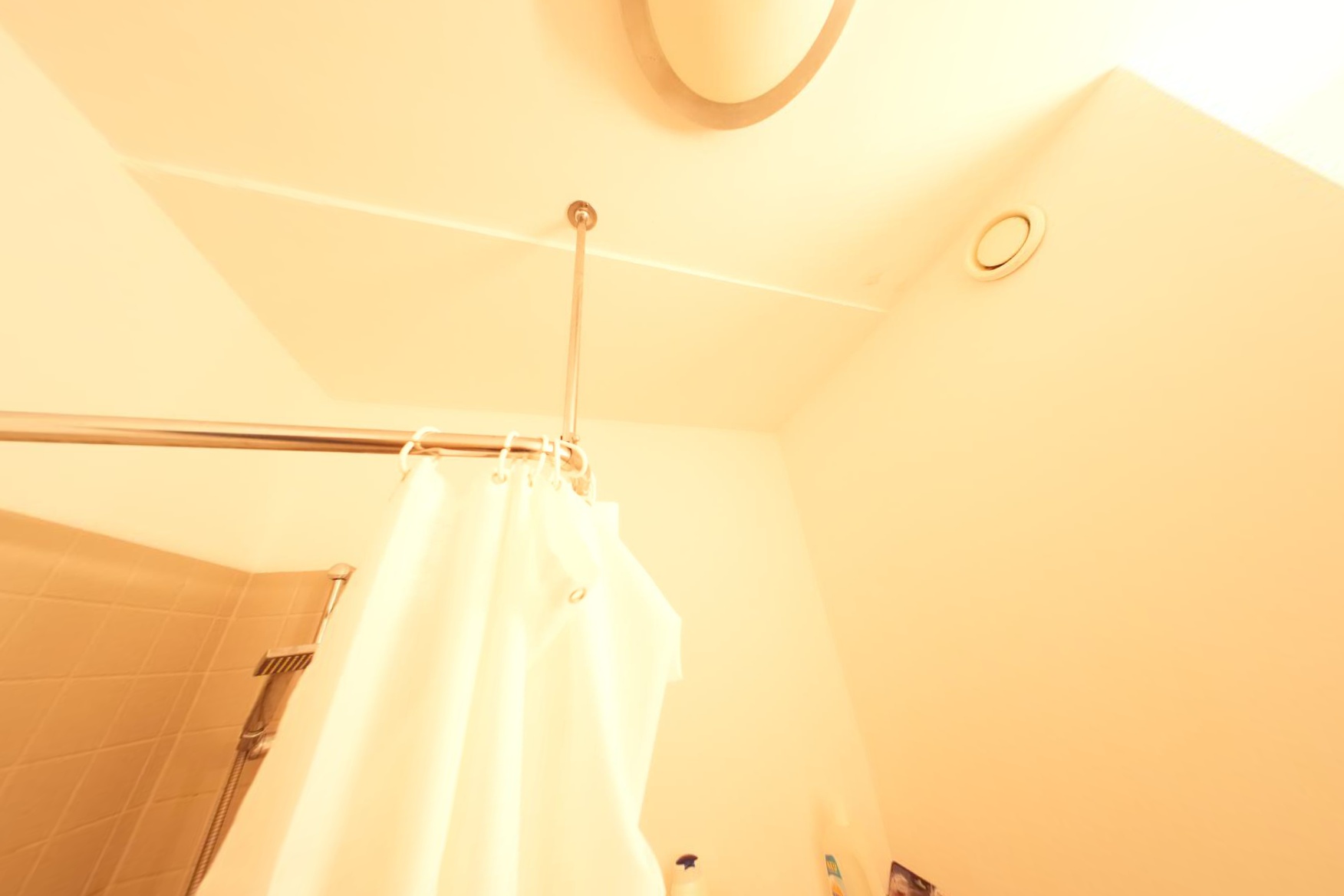} &
        \includegraphics[width=0.22\textwidth]{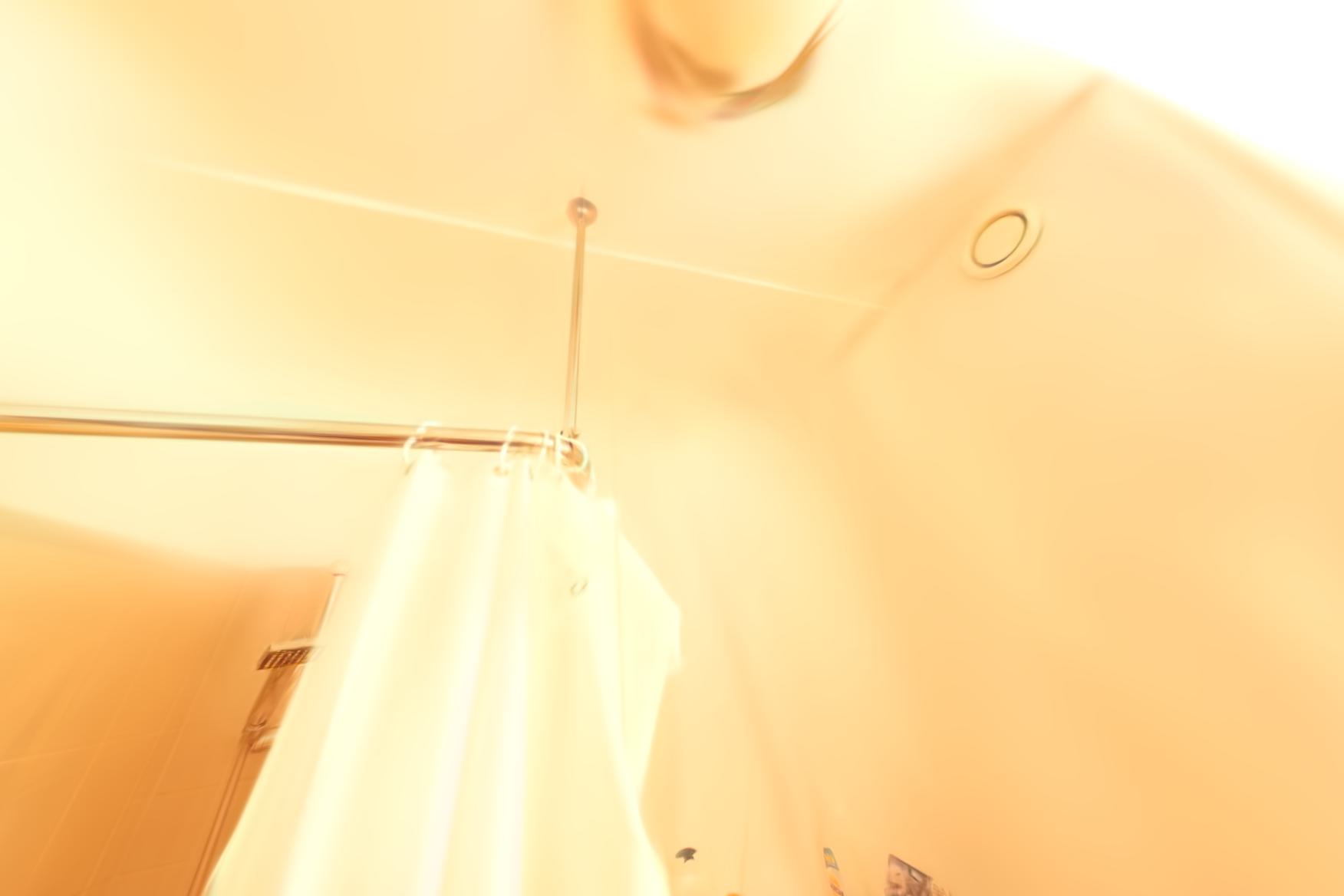} \\ 

        & \multicolumn{2}{c}{\multirow{2}{*}{Source views}} & \multirow{2}{*}{Test image} & Rendered from  \\
        &                        &                           &                             & estimated pose \\
    \end{tabular}
    \caption{\textbf{ScanNet++.} Examples of qualitative results obtained by our cross-scene model on diverse scenes of ScanNet++ \cite{yeshwanthliu2023scannetpp}. From left to right, two images from the training images, a test query image, and the rendered image from our model pose prediction.}
    \label{fig:scannet_examples}
\end{figure*}

Below, we provide additional details about our implementation and examples of challenging scenes from the ScanNet++ dataset in Figure \ref{fig:scannet_examples}. As shown, the scenes are large and diverse, including extensive texture-less areas, which demonstrates the generalization ability of our cross-scene model.

Our code is still a work in progress. We will publish it after finishing to clean and refactor it for easy use. All of the data we used is publicly available. We will also release our pre-trained models.

\section{Gaussain Splatting Implementation Details}
\label{sec:supp_gs}
We use the pre-built COLMAP reconstructions from \cite{brachmann2021limits} for the 7scenes dataset and the reconstructions provided in HLoc toolbox \cite{sarlin2019coarse} for the Cambridge landmarks dataset. We train all the scenes using the vanilla 3DGS \cite{kerbl3Dgaussians}, for 30k iterations using the default parameters, in Table \ref{tab:gs_psnr} we report the per-scene PSNR scores for our trained models on the training images. Notably, the rendering quality of outdoor scenes is inferior compared to indoor scenes, which might explain the degradation in our pose estimation accuracy.

\PAR{Handling challenges in outdoor scenes.}To effectively train a 3DGS model for outdoor scene reconstruction, we focus on reconstructing static elements such as buildings, fences, and signs. This approach addresses real-world challenges like varying lighting conditions, dynamic objects, and distant regions. To mitigate these issues, we use a pre-trained semantic segmentation model~\cite{cheng2022masked} to mask out sky regions and moving objects, including pedestrians and vehicles. These elements, which constitute only a small portion of the captured images, are excluded from the loss function during training, resulting in more accurate scene reconstruction. For this purpose, we utilize pre-computed segmentation maps provided by \cite{zhou2024nerfect}, generated using the method described in \cite{cheng2022masked}.

\end{document}